# Data-based wind disaster climate identification algorithm and extreme wind speed prediction

Wei Cui[a]; Teng Ma[b]\*; Lin Zhao[c,d]\*; Yaojun Ge[c,d]

[a] Assistant Professor, State Key Lab of Disaster Reduction in Civil Engineering, Tongji University, Shanghai 200092, China;

[b] Undergraduate student, College of Civil Engineering, Tongji University, Shanghai 200092, China; Corresponding author, E-mail address: tengma_725@163.com

[c] Professor, State Key Lab of Disaster Reduction in Civil Engineering, Tongji University, Shanghai 200092, China; Corresponding author. Tel.: +86-135-6469-6598; Fax: +86-21-6598-4882, E-mail address: zhaolin@tongji.edu.cn

[d] Professor, Key Laboratory of Transport Industry of Wind Resistant Technology for Bridge Structures, Tongji University, Shanghai 200092, China

## Abstract

An extreme wind speed estimation method that considers wind hazard climate types is critical for design wind load calculation for building structures affected by mixed climates. However, it is very difficult to obtain wind hazard climate types from meteorological data records, because they restrict the application of extreme wind speed estimation in mixed climates. This paper first proposes a wind hazard type identification algorithm based on a numerical pattern recognition method that utilizes feature extraction and generalization. Next, it compares six commonly used machine learning models using K-fold cross-validation. Finally, it takes meteorological data from three locations near the southeast coast of China as examples to examine the algorithm's performance. Based on classification results, the extreme wind speeds calculated based on mixed wind hazard types is compared with those obtained from conventional methods, and the effects on structural design for different return periods are discussed.

## Keywords

Extreme wind speed; Mixed climates; Data-driven method; Pattern Recognition; Machine Learning;

## 1. Introduction

Wind effects are key factors in structural design, and extreme wind speeds are the starting point. For flexible structures such as long-span bridges, long-span roofs and high-rise buildings, wind loads are normally the predominant loads. In order to meet both the ultimate safety and performance requirements of wind-resistant structural design, it`s necessary to accurately estimate the extreme wind speeds for different recurrence periods.

For significant buildings and infrastructures, it is necessary to estimate the extreme wind speed through probabilistic methods from local wind speed records. The probabilistic method and extreme value theory have already been extensively applied in both building design and structural wind engineering research. The most widely used method of extreme wind speed estimation includes three steps: wind speed sample extraction, probabilistic distribution model selection and model parameter fitting. There are three major methods for wind speed sample selection: the stage extremum method, the peak-over-threshold method and the method of independent storm (Palutikof, Brabson et al. 1999). Through sampling extreme values by extracting peak wind speeds from unit time intervals, the stage extremum method uses the Gumbel distribution (Gumbel 2012) to fit the extreme wind speed distribution and estimates design wind speeds for different return periods. This method is easy to implement but has low sample utilization rate, so it is suitable for areas with long-term wind speed observation data. It is widely used by design codes in different countries, including the Canadian Building Structure

Load Specification (CNBC1995) and the Chinese Building Load Specification (GB50009-2012). In order to overcome the problem of low utilization rate of wind speed observation data, Cook (1982) proposed a new method for selecting wind speed samples, namely, the method of independent storm (MIS). Several later studies (Harris 1999) have shown that if the wind speed observation data comprises continuous data samples and an independent storm segment can be identified, the MIS proposed by Cook has a better data utilization rate than the stage extreme value method. The peak-over-threshold method filters data samples below a pre-defined threshold from the parent sample as an extreme value sample. In 1975, Pickands (1975) first proposed this extreme value theory based on the Generalized Pareto distribution. In 1996, Simiu and Heckert (1996) used this method to analyze extreme wind speeds in the United States. In all the above three methods, it is assumed that the wind speed data are sampled from the same probabilistic distribution, which means wind extremums comes from the same climate pattern, such as monsoons.

However, in mixed climate regions affected by various wind disasters, such as China's southeastern coastal region, this assumption is invalid. Samples with the same wind speed values but different wind hazard climate types may yield different extreme wind speed estimations. If a single distribution is used for fitting, relatively large deviations will occur. Based on the probability distribution characteristics of wind speeds from mixed climate types, (Gomes and Vickery 1976, Gomes and Vickery 1978) proposed a composite extreme wind speed analysis method for mixed climates including thunderstorms, hurricanes and tornadoes, and carried out extreme wind speed estimations for various return periods. Subsequently, Cook and Harris (2003) and Cook (2004) improved the method by including the confidence interval analysis of wind speeds with mixed wind climates. The proposed skewed Gaussian distribution is suitable for describing the wind pressure distribution on a building and the error is weakened when fitting the tail part of the Gomes-Vickery method (Cook, Ian Harris et al. 2003, Cook 2004). In addition, many other scholars have developed many alternative methods for distinguishing between thunderstorm and non-thunderstorm climates, and combined their probability distributions into mixed distributions for extreme analysis (Riera and Nanni 1989, Twisdale and Vickery 1992, Choi 1999, Choi and Hidayat 2002). In summary, the composite extreme wind speed analysis method is suitable for regions with mixed climates.

However, most meteorological observatories only record conventional wind climate data such as wind direction and wind speed, and do not record the associated wind climate categories, which makes screening various wind climate types very difficult and tedious. This limitation also constrains research, application and development of extreme wind speed prediction with mixed climates.

Different methods and benchmark procedures have been proposed to identify different types of wind disasters from conventional meteorological data. Riera et al. (1989) extracted thunderstorm fragments from conventional meteorological data based on the duration of thunderstorms, the occurrence of lightning, and rainfall (Riera and Nanni 1989). Choi et al. divided wind disasters in Singapore into large-scale and small-scale wind disasters by visually observing the meteorological data (Choi and Tanurdjaja 2002). Durañona et al. (2007) proposed four indicators to discriminate among extreme winds caused by different wind climates based on extreme wind speed, average wind speed and wind speed trend (Durañona, Sterling et al. 2007). Lombardo et al. (2009) achieved an automatic separation of thunderstorms and non-thunderstorms based on the start and end times of thunderstorms, and pointed out that the conditions for ensuring the independence of the two extreme wind speed data points are different (Lombardo, Main et al. 2009). Thunderstorms that need to be separated should have a 4-hour sampling rate rather than a 4-day one. The above methods can achieve rapid identification and extraction of wind disaster segments, and the above algorithms are based on empirical criteria, which are obtained from raw data analyses. It is difficult to unify judging criteria through different regions; one method is only applicable to the identification of certain wind disasters in a specific area. Therefore, it is difficult to propose a universal wind disaster identification method based on traditional experience-based criteria.

This paper tries to break through the limitations of traditional experience-and-history-based identification methods, and proposes a data-based wind disaster type smart identification solution. In order to maximize utilization of wind disaster data for wind disaster identification, machine learning algorithms are used to extract wind disaster data features that have high correlation with the wind disaster type from structured conventional wind climate data, and thus achieve automatic identification of wind disaster types. This paper introduces three example stations—Zhoushan, Dachen Island, and Zhangzhou—to verify the algorithm's efficiency and accuracy based on single meteorological station history data identification and cross-identification of cross-station in the same area. Based on identified wind disaster data, extreme wind speed calculation results for different return periods of mixed distributions are compared with those obtained from traditional methods to verify the proposed method for extreme wind speed prediction in a mixed climate area. Finally, the applicability and limitations of the automatic identification algorithm for wind disaster types are discussed and the main conclusions of this paper are summarized.

## 2. Data Source and Preprocessing

This section introduces the original data sources and filtering, wind disaster segmentation and other data preprocessing steps in detail. Related methods of wind speed time series and surface roughness modification are also introduced in detail in the fourth chapter of the actual example because of the strong geographical relevance to meteorological observatories.

### 2.1. Meteorological data source

The original meteorological data originate from the NOAA Global Integrated Surface Database. This database consists of observations from 29,570 meteorological stations around the world, and includes wind direction, wind speed, atmospheric temperature, atmospheric humidity, atmospheric pressure, precipitation, visibility, cloud conditions, etc.

The region studied in the paper is China's southeastern coast, which is affected by multiple wind climates including East Asian monsoons, Northwest Pacific typhoons and other microscale winds. Therefore, the main wind damage types to be identified in the paper are monsoon type, typhoon type and other types. The other types are defined as a continuous process with high wind speed but with no typical typhoon or monsoon characteristics.

### 2.2. Wind disaster segmentation

The proposed data separation algorithm for extreme wind speeds aims to classify the wind climate type from raw data from meteorological observatories. Before the algorithm is formally presented, because the raw data includes many variables including wind, precipitation, cloud, temperature, etc., it is firstly necessary to select relevant ones. Five time series of data with high correlation with wind damage types, wind direction, wind speed, air pressure, temperature and precipitation, are extracted from the database as original data.

A wind disaster represented by meteorological data is a continuous process rather than a single moment. Taking typhoons and monsoons as examples, these two common wind damage types would normally last several hours or even days. Therefore, before identifying the wind disaster type, it is necessary to select multidimensional time series segments with high wind speeds from several years of meteorological records.

Data segments with maximum wind speeds greater than 12m/s within a set period of time are extracted from original raw data. The wind-range frequency spectrum of the natural wind compiled by Van der Hoven (1957) shows that the macro-meteorological peak is at a center period of about 4 days, which means a wind process in a natural weather system seldom lasts longer than 96 hours (Van der Hoven 1957).

Therefore, we chose a time series with a span of 96 hours as the initial wind disaster segment with a time step with maximum speed at the center of the 96-hour interval. The extracted initial time series data fragments covered all wind disaster durations. However, normally some redundant low speeds records are also included because wind disaster climates last less than 96 hours.

In order to avoid the influence of low-speed data on the wind damage classification model, the heuristic segmentation algorithm (BG algorithm) proposed in Ivanov et al. (2001) is employed to remove it from the initial wind damage fragments (Bernaola-Galván, Ivanov et al. 2001). This algorithm recursively divides the time series into subsequences. Test statistical values of each point in the time series are calculated and the maximum is taken as the split point. Each subsequence has two recursive terminating conditions: that the statistical significance of the maximum test statistic value is less than critical value $P_0$ and that the length of the subsequence is less than the minimum segmentation scale $l_0$.

Through this operation, the initial wind disaster segment can be divided into multiple sub-segments, and the sub-segment including the maximum wind speeds at the middle time point is taken as the final windstorm segment. Considering the duration of the windstorm and the actual segmentation effect, we take $P_0 = 0.7$ and $l_0 = 8$ to segment the initial windstorm segment.

Figure 1 shows an example of the wind disaster process. The example wind speed data started from January 9th, 2010 at Zhoushan, China. The first segmentation divides the original time series into a blue line (non-storm segment) and a red line (storm segment) based on maximum wind speed at center point. Then, using the BG algorithm, high- and low-wind-speed subsequences are identified. The test statistical values are shown as a purple dotted line and the BG split point is at -3h, which is the maximum test statistic. The red line is divided into a solid line and a dashed line by BG split point. In this case, the storm segment only has one recursion. Finally, the solid line is used to present this whole storm process.

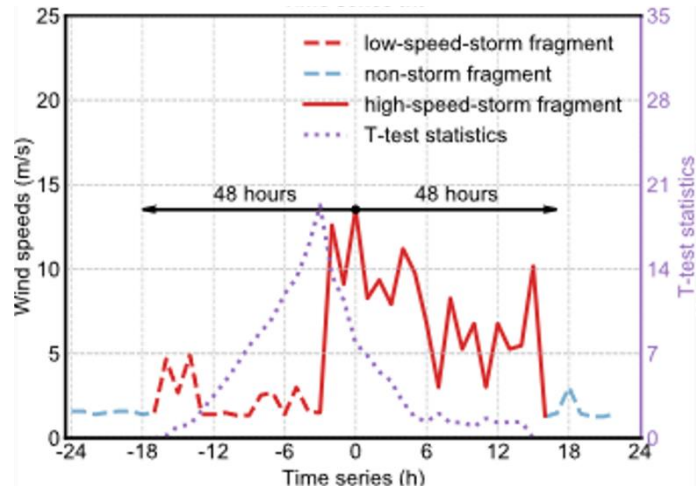

Figure 1 Wind disaster fragment screening and segmentation diagram

## 2.3. Wind disaster sample correlation

An important precondition of extreme analysis is that the data are statistically independent, so each wind disaster fragment can only be represented by one wind speed. Each wind disaster fragment obtained through the proposed algorithm consists of the maximum wind speed occurring every three hours. The wind disaster usually lasts from several hours to several days, so generally several extreme wind speeds can be obtained from each wind disaster fragment. Therefore, in order to ensure the independence of extreme wind speed samples, an appropriate method is required to extract the maximum wind speed from each wind disaster and eliminate other wind speeds related to the same wind disaster. Simiu et al. (Simiu and Heckert 1996) believed that sample independence can be guaranteed by the length of the time interval between adjacent extreme sample points, that is, the time interval should be greater than the duration of typical wind disasters.

The method for constructing extreme data sets also uses the method of choosing a time interval that ensures the independence of extreme data sets. In the wind disaster filtering and segmenting method introduced in the second section, the method of peak-over-threshold is used to preliminarily filter, which ensures that the interval between the maximum wind speed points of the wind disaster segments are over 96 hours. Segmenting by the BG method ensures that each wind disaster fragment belongs to an independent wind disaster system. Therefore, each wind disaster is represented by the maximum wind speed selected from it. That is to say, an independent wind speed sample with reduced statistical dependence containing information of types is constructed.

## 3. Establishment of Wind Disaster identification Algorithms

In the data science field, machine learning algorithms can find regulations just from data sets, and then make predictions. This is the major difference between machine learning data-based models and traditional knowledge-based models. The two common machine learning algorithm categories are supervised learning and unsupervised learning depending on whether the data set contains classification label information. As an initial study on wind climate classification, the proposed machine learning algorithm is a supervised learning algorithm, which means models should comprise labeled data before implementation of the proposed learning model.

### 3.1. Feature Extraction

Typically, the classification of time series is based on the features rather than the actual data value in time series (Bishop 2006). The classifier algorithm is very sensitive to data errors, because data at each time step contributes equally to the classification results, and errors accumulated through the whole time series have a relatively big impact on the model's accuracy (Witten, Frank et al. 2016). In addition, in model training and identification, data vectors are required to have the same length. However, wind disaster data fragments to be classified have different time lengths because of the BG algorithm. Therefore, an appropriate wind disaster feature extraction method is required.

In order to distinguish wind disaster types from meteorological data, wind speed, wind direction, atmosphere pressure, air temperature and precipitation are used by the feature extraction method, which is based on statistical results. According to (Xi, Keogh et al. 2006), the segmented low-frequency multi-dimensional time series data can be represented by 8 common statistical features for each time series data: mean value $\mu$, standard deviation $\sigma$, skewness $\lambda$, kurtosis $\kappa$, maximum value, minimum value, range value and median value. Skewness and kurtosis contain information on the shape of the distribution of the time-series values. More precisely, skewness characterizes the degree of asymmetry of values around the mean value. Kurtosis measures the relative peakedness or aptness of the value distribution relative to a normal distribution.

The feature extraction methods for wind disaster data in this paper can be separated into three types: first-order features, second-order features and environmental features (Nanopoulos, Alcock et al. 2001). First-order features are based on the actual values of the series $x(t)$ and second-order features are based on the differences between nearby values $x'(t)$ that contain the original series varying trends, and can also be used to filter noise. Environmental features include geographic location and wind disaster occurrence months, which are not included in original data, but are helpful for model classification.

## 3.2. Data Set Establishment

In this paper, the data set of the wind disaster identification algorithm is constructed by a manual classification method. According to the characteristics of wind speed, wind direction, rainfall, temperature, air pressure and typhoon history database, wind disaster data fragments are classified manually (Cook, Ian Harris et al. 2003). According to the geographical climate features, monsoon and typhoon are the usual wind disasters in the coastal area of southeastern China. The feature of a monsoon wind disaster is that the wind direction fluctuates around a certain value in a small variation range, while the feature of a typhoon wind disaster is that before the typhoon lands, air pressure decreases and wind speed increases, and a large amount of precipitation is normally recorded concurrently. After the typhoon's passing, air pressure rises and wind deceases.

In order to reduce the probability of mislabeling and omission of typhoon manual annotation, the typhoon database is used for cross-validation. If a segment of a wind disaster does not appear in the typhoon database, the wind disaster belongs to another type.

## 4. Application Examples

In order to present the building and validation of a machine learning model, this section takes Dinghai meteorological station as an example and compares six widely-used machine learning algorithms. Two performance indices, confusion matrix (Batista, Prati et al. 2004) and ROC curve (Bradley 1997), are introduced to evaluate the identification effectiveness of each wind disaster identification algorithm. The one with the best performance is then proposed as the wind disaster classification algorithm, and the other two meteorological stations, Dachen island and Shengzhou, are employed as application cases. The flow chart of the wind disaster classification algorithm is shown in Figure 4.

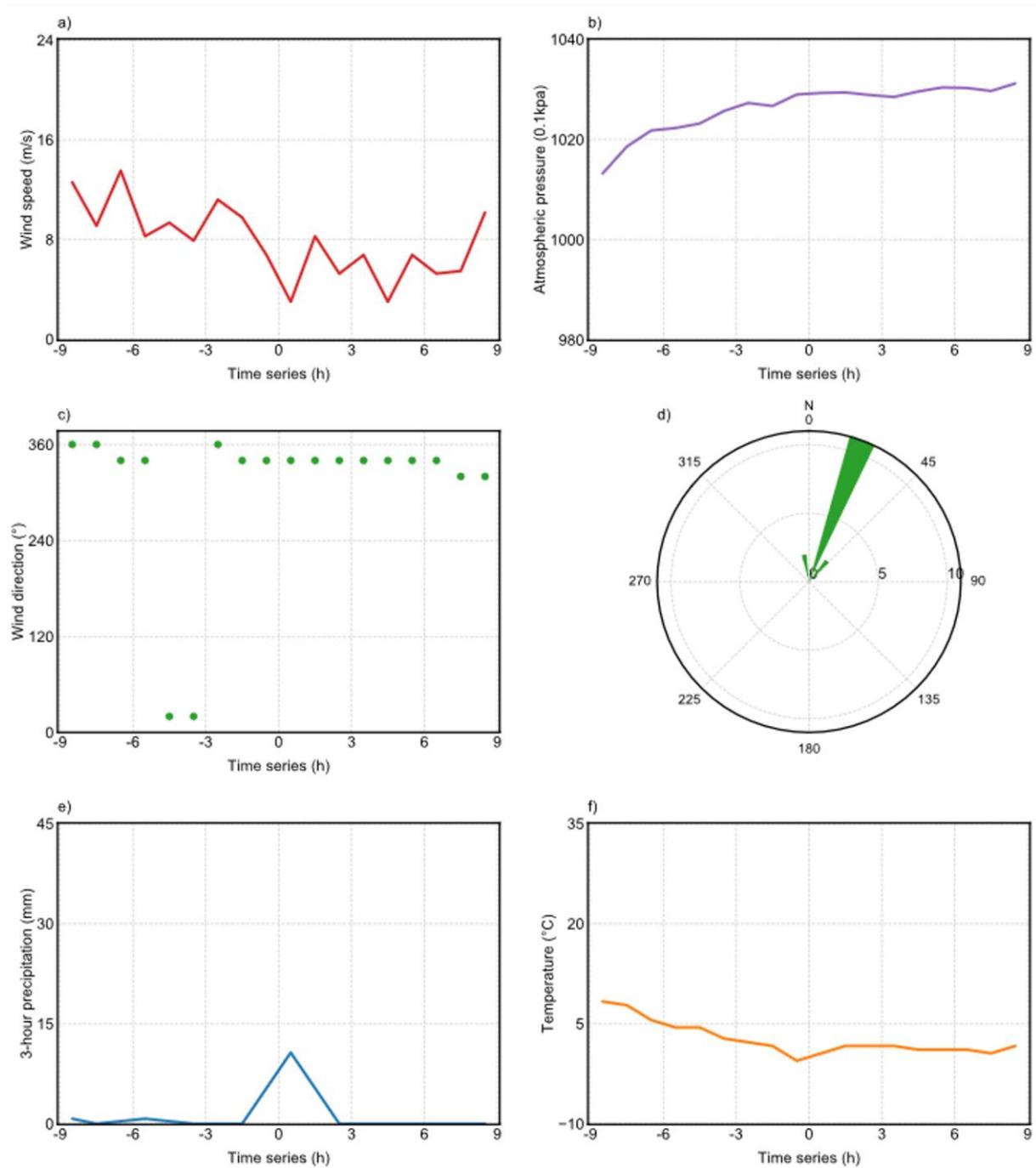

Figure 2　Monsoon-type wind disaster data chart start from March 10th, 2005 in Dinghai meteorological station: a) wind speed time-history, b) atmosphere pressure time-history, c) wind direction time-history, d) polar histogram of the wind direction, e) precipitation time-history and f) temperature time-history

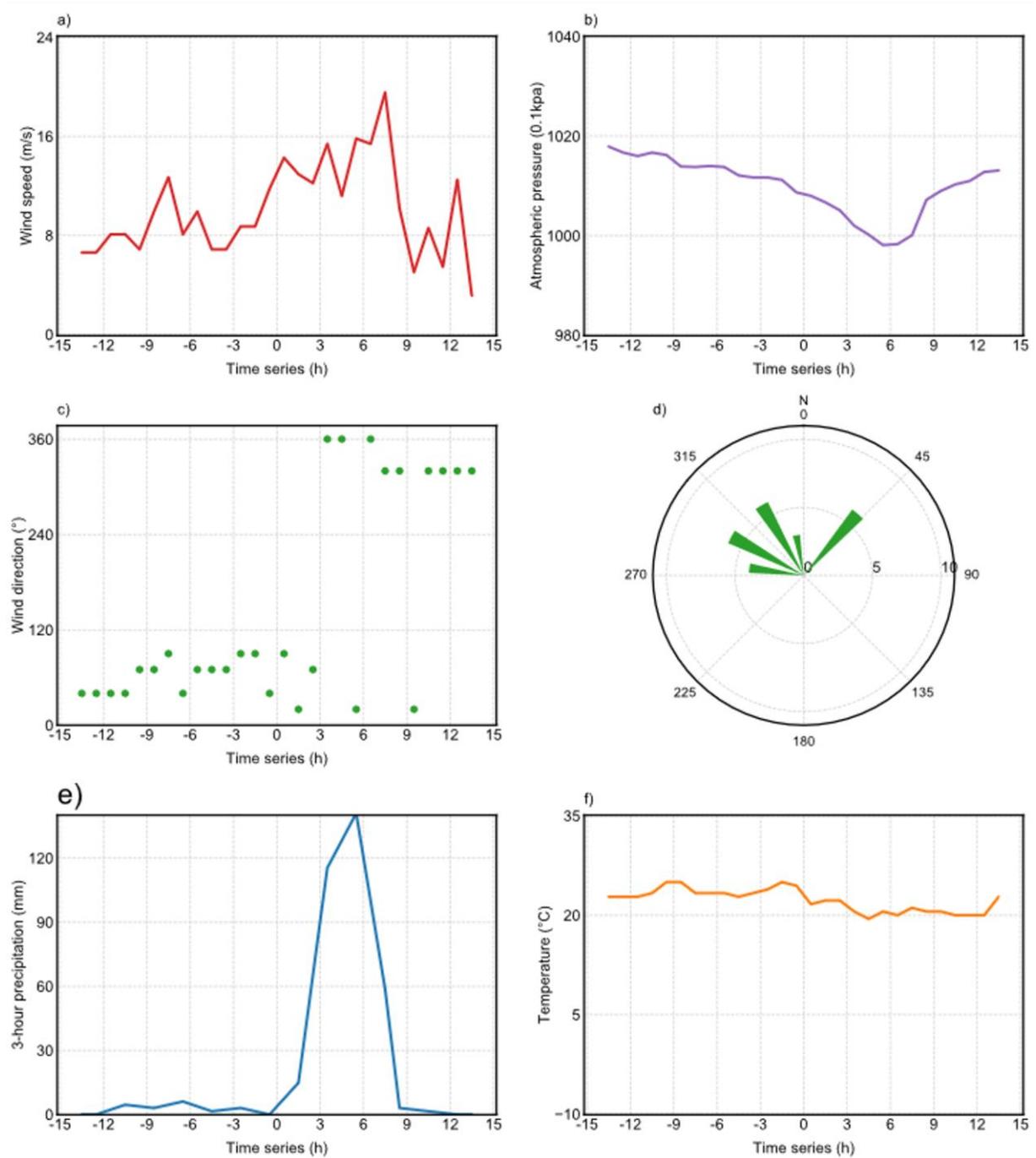

Figure 3  Typhoon Seth (1994) wind disaster data chart starting from October 8th, 1994 in Dinghai meteorological station: a) wind speed time-history, b) atmospheric pressure time-history, c) wind direction time-history, d) polar histogram of the wind direction, e) precipitation time-history and f) temperature time-history

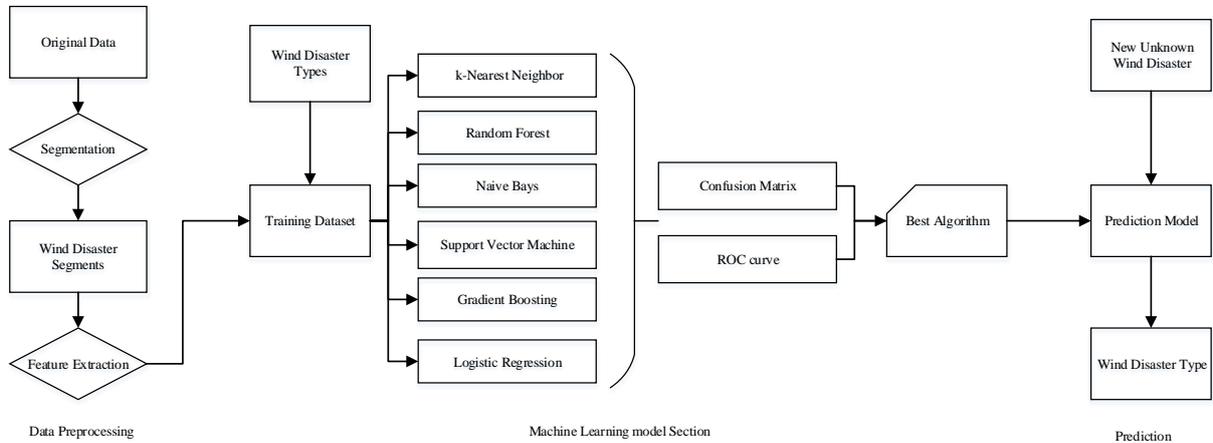

Figure 4 Wind disaster classification algorithm flow chart

## 4.1. Introduction of meteorological stations

Three meteorological stations, Dinghai, Dachen Island and Shengzhou, are employed in this study. Figure 5 shows the geographical location and terrain around each station. They are all located near the coastline of the East China Sea and are within 200 kilometers of each other. It can be considered that they are all affected by the same climate pattern, and have the same wind environment. However, the geomorphological profiles around them are different. Dinghai Station is located in a coastal city and has many buildings and urban infrastructures. Dachen Island Station is an off-shore island and is surrounded by the ocean. Shengzhou Station is in an internal city about 70 km from the coast. Differences in geomorphology influence the data of these stations, making them suitable for the model performance examination.

## 4.2. Dataset Construction

The original meteorological data from Dinghai, Dachen Island and Shengzhou stations all have

27 years of records from 1990 to 2016, with total valid data of 78105, 77972 and 71696 data

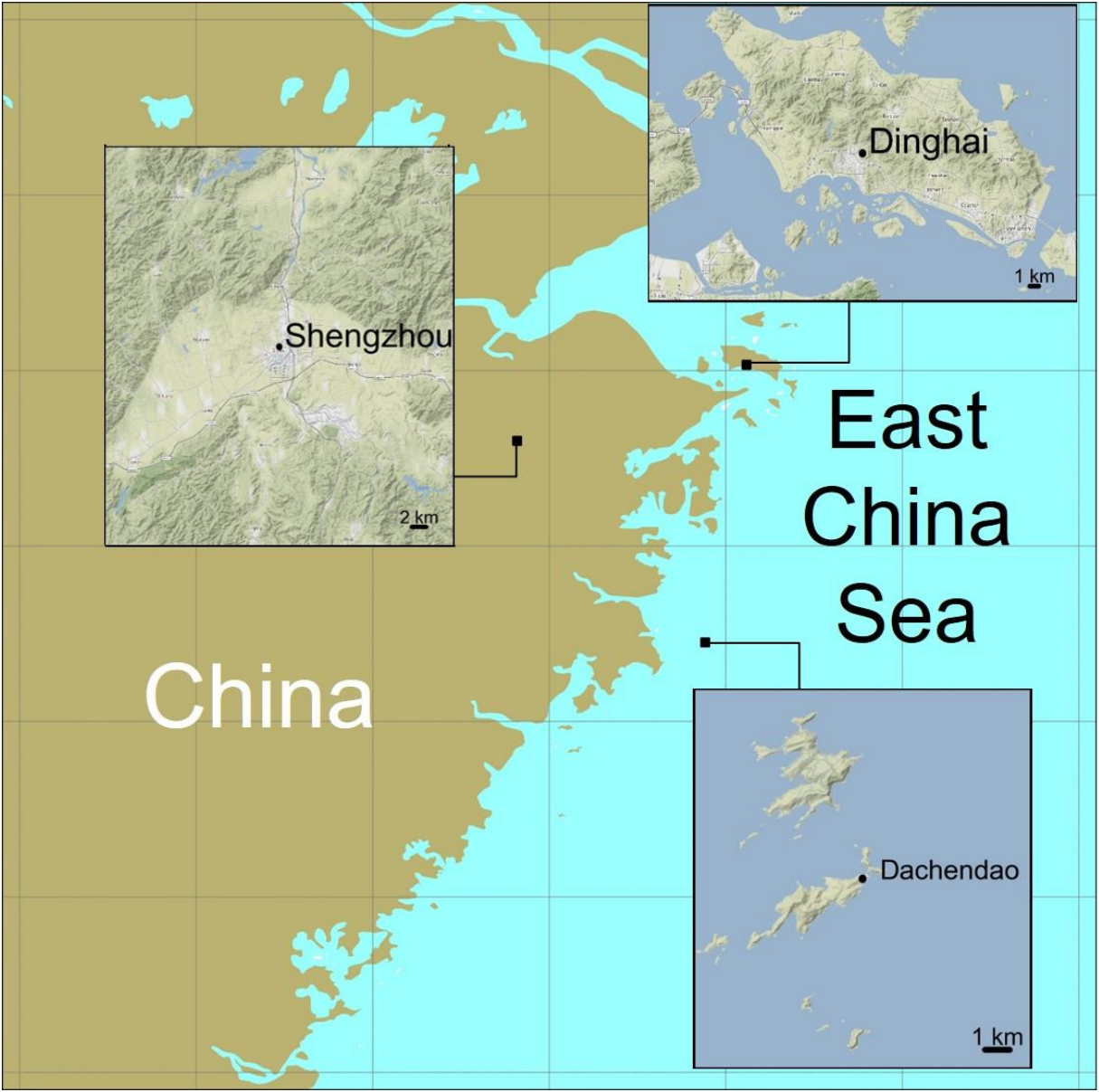

Figure 5 Geographical observation station geographical locations and topographic map

points, respectively. Before the extraction of wind disaster fragments, in order to consider terrain interference effects around each station, ground roughness modification to the original wind speed records will be performed. All wind speeds from the three stations will be transformed to uniform ground roughness: 10 m above open terrain ($z_0 = 0.03$ m) (ESDU,

1984). The transforming factors are shown in Table 1.

After getting the time-histories of wind speed data with standardized ground roughness, the threshold is set to extract the wind disaster data fragments from the original data. The wind speed threshold is set to 12 m/s (Simiu and Heckert 1996) to preprocess the data according to the method of section 2, and 678, 730 and 289 wind disaster fragments are obtained for Dinghai, Dachen Island and Shengzhou stations. After manually labelling the wind disaster types, the three types, typhoon, monsoon and other, histograms for each station are derived as shown in Figure 6.

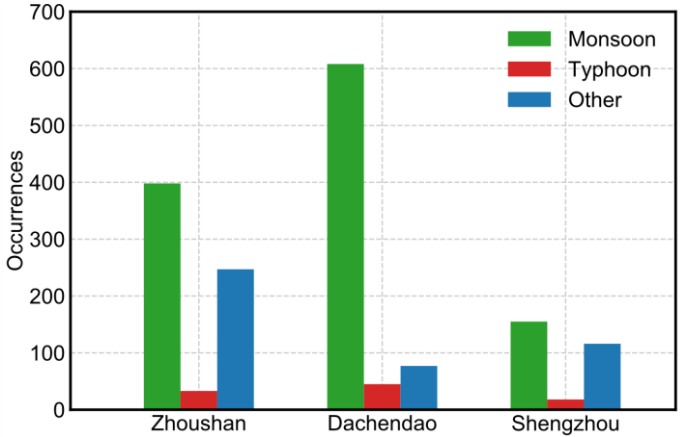

Figure 6 Statistical histogram of wind disaster data fragments of example station

Table 1. Transforming Factor of Roughness of Different Wind Directions for Dachen Island, Dinghai, Shengzhou Meteorological Station Data

| Station Name | Wind Direction | Correction Factor | Wind Direction | Correction Factor | Wind Direction | Correction Factor |
|---|---|---|---|---|---|---|
| Dachen Island | 30 | 1.029 | 150 | 1.035 | 270 | 0.943 |
| | 60 | 1.093 | 180 | 1.024 | 300 | 1.035 |
| | 90 | 1.052 | 210 | 1.012 | 330 | 1.087 |
| | 120 | 1.903 | 240 | 1.018 | 360 | 1.058 |
| Dinhai | 30 | 0.895 | 150 | 0.900 | 270 | 0.820 |
| | 60 | 0.950 | 180 | 0.890 | 300 | 0.900 |
| | 90 | 0.915 | 210 | 0.880 | 330 | 0.945 |
| | 120 | 0.905 | 240 | 0.885 | 360 | 0.920 |
| Shengzhou | 30 | 0.895 | 150 | 0.900 | 270 | 0.820 |
| | 60 | 0.950 | 180 | 0.890 | 300 | 0.900 |
| | 90 | 0.915 | 210 | 0.880 | 330 | 0.945 |
| | 120 | 0.905 | 240 | 0.885 | 360 | 0.920 |

## 4.3. Application of Machine Learning Algorithms and performance comparison

According to the feature extraction method described in Chapter 3.1, the machine learning model will accept 82-dimensional features as input, including 40-dimensional first-order features, 40-dimensional second-order features and 2-dimensional environment features. The model output is wind disaster type: typhoon, monsoon and other.

In this paper, six commonly used machine learning algorithms, k-Nearest Neighbor (KNN), Naive Bays (NB) (Domingos and Pazzani 1997), Support Vector Machine (SVM) (Cortes and Vapnik 1995), Gradient Boosting (GBDT) (Friedman 2001), Random Forest (RF) (Breiman 2001) and Logistic Regression (LR) (Press and Wilson 1978) are selected for fitting test to determine the most suitable algorithm model for learning wind disaster data fragment information.

The K-fold cross validation test method is utilized to compare the performances of these six algorithms. The $K$-fold cross validation test method (Kohavi 1995) divides the data set into $K$ groups. Each sub-dataset is used as a testing set once, and the other $K - 1$ sub-datasets are used as training sets to get $K$ testing-training dataset combinations, as shown in Figure 7.

Mean value and variance of the prediction results obtained from the $K$ models are used as the evaluation parameters of the classifier. K-fold cross validation can effectively test the over-fitting and under-fitting states (Kohavi 1995), and verify the model's accuracy, reliability and robustness. In this paper, $K$ is chosen as 10, and the algorithm's performance for cross validation, which is defined as accuracy in subsection 4.4, is shown in Table 2 and Figure 8. It can be seen that the accuracy of SVM, LR, KNN and random forest are all greater than 0.75, and the accuracy variance of SVM is the smallest, which indicates that the generalization ability of the SVM model is the highest. Therefore, the SVM support vector machine algorithm is finally chosen to realize the wind disaster identification model.

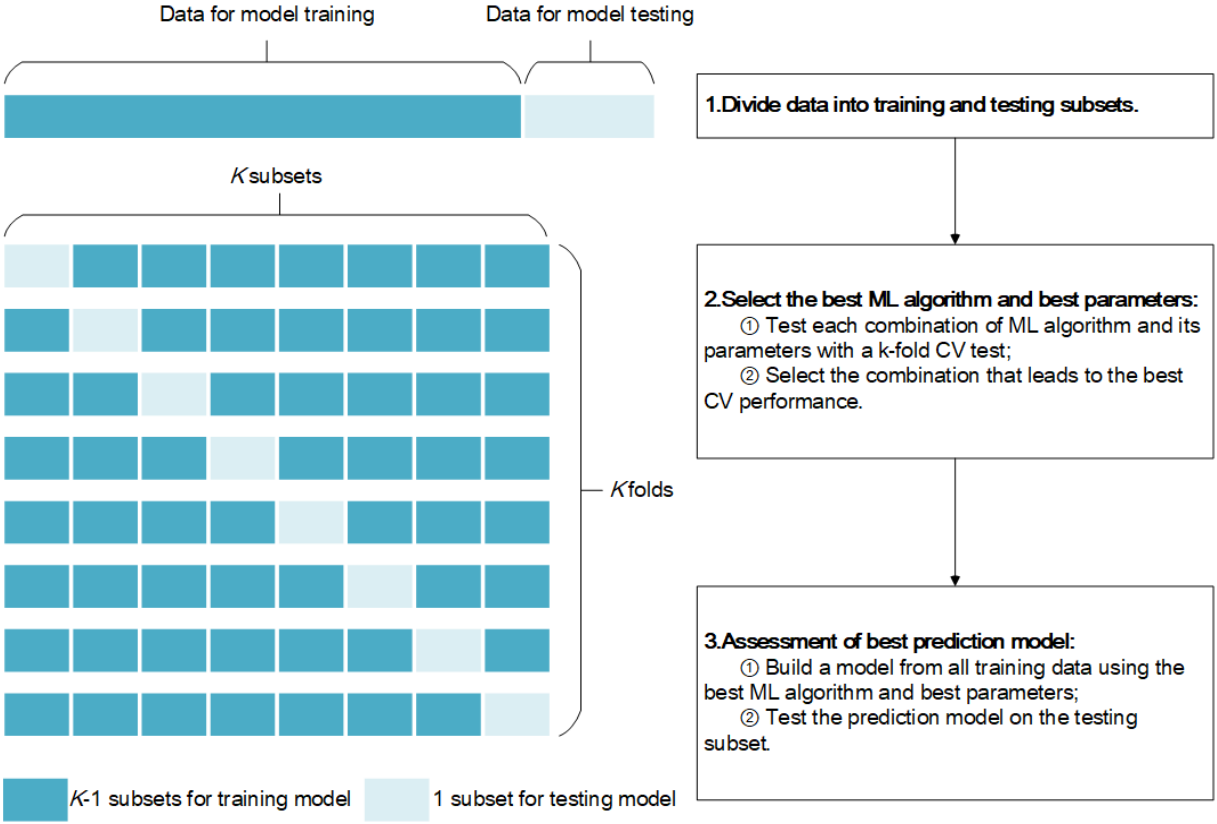

Figure 7. K-fold cross-validation method diagram

Table 2. 10-fold cross-validation accuracy rate statistics for six machine learning models

| Algorithm | Mean of accuracy | Accuracy standard deviation |
|---|---|---|
| k-Nearest Neighbor (KNN) | 0.7906 | 0.0694 |
| Naive Bayes (NB) | 0.7677 | 0.0738 |
| Support Vector Machine (SVM) | 0.8064 | 0.0414 |
| Gradient Boosting (GBDT) | 0.7841 | 0.0716 |
| Random Forest (RF) | 0.7483 | 0.0398 |
| Logistic Regression (LR) | 0.7794 | 0.0842 |

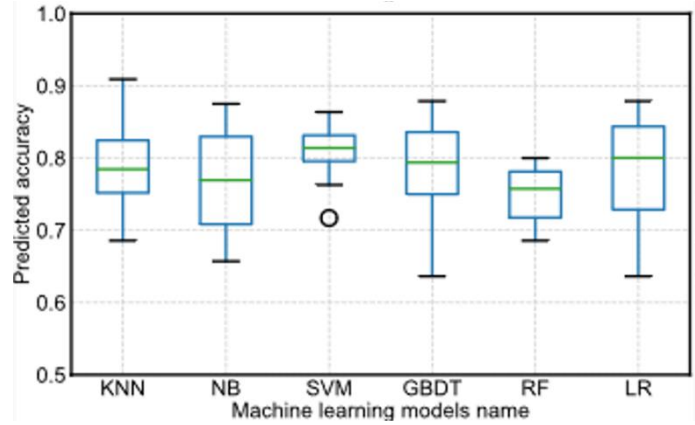

Figure 8. 10-fold cross-validation accuracy box plot for six machine learning models

## 4.4. Single Meteorological Station History Data Identification

After determining the best machine learning algorithm, the model's performance should be evaluated. Each station's data is divided into training datasets and test datasets in the proportion of 7:3, which means each station uses 70% of the data to train the model and 30% of the data to test the model. The concepts of confusion matrix, precision, recall and average accuracy are introduced to quantify the evaluation performance. The classification results can be divided into four categories according to the model's prediction results and the actual types of wind disaster data, as shown in Table 3.

Table 3. Second Classification Problem Confusion Matrix

| | | Actual Value (as manually labelled) | | Index Formula |
|---|---|---|---|---|
| | | positives | negatives | $\text{TP rate} = \dfrac{\text{TP}}{\text{TP} + \text{TN}}$ |
| Predicted Value (predicted by model) | positives | **TP** true positive | **FP** false positive | $\text{FP rate} = \dfrac{\text{TP}}{\text{FP} + \text{FN}}$ |
| | | | | $\text{Precision} = \dfrac{\text{TP}}{\text{TP} + \text{FP}}$ |
| | negatives | **FN** true negative | **TN** true negative | $\text{Recall} = \dfrac{\text{TP}}{\text{TP} + \text{FN}}$ |

In Table 3, predicted value means wind disaster category judged by the model through wind disaster characteristics and actual value means manually labeled windstorm category. The true positive (TP) samples represent the number of samples that are predicted to be positive and actually labeled as positive in all data. The false negative (FN) samples represent the number of examples that are predicted to be false and actually labeled as positive in all data. The true negative (FP) samples represent the number of examples that are predicted to be positive and actually labeled as false in all data. The true negative samples (TN) represent the number of examples that are predicted to be false and actually labeled as false in all data. On this basis, performance measurement indexes such as Precision and Recall are defined. Precision represents the proportion of real cases to all positive predictions, which is equivalent to the degree of accuracy of the measurement model identifying positive samples. Recall represents the proportion of true positive cases to all actual positive cases, which is equivalent to the degree of coverage of the measurement model identifying positive samples. The accuracies of the wind disaster identification models from the three stations are shown in Table 4.

Table 4. Dachen Island, Dinghai and Shengzhou Meteorological Station Data Set Identification Results of Various Types of Wind Disasters

| Station | Wind disaster type | Precision | Recall |
|---------|-------------------|-----------|--------|
| Dachen Island | **Typhoon** | 0.889 | 0.615 |
| | **monsoon** | 0.905 | 0.973 |
| | **Other** | 0.810 | 0.548 |
| Dinghai | **Typhoon** | 1.000 | 0.765 |
| | **monsoon** | 0.870 | 0.811 |
| | **Other** | 0.758 | 0.887 |
| Shengzhou | **Typhoon** | 1.000 | 0.833 |
| | **monsoon** | 1.000 | 1.000 |
| | **Other** | 0.678 | 0.379 |

On this basis, all samples are ranked according to a confidence coefficient from high to low, and the current False Positive Rate and True Positive Rate are calculated one by one as positive samples and plotted as a receiver operating characteristic (ROC) curve in Figure 10. The ROC curve is a comprehensive indicator reflecting the continuous variables of sensitivity and specificity. It is a compositional method that reveals the relationship between sensitivity and specificity. The area between the ROC and the x-axis is called "average accuracy", which corresponds to the overall simultaneous measurement of the model's accuracy and recall. The average accuracy is a widely-used model performance index. The ROC of the wind disaster identification models for the three stations are shown in Figure 9, Figure 10 and Figure 11, respectively.

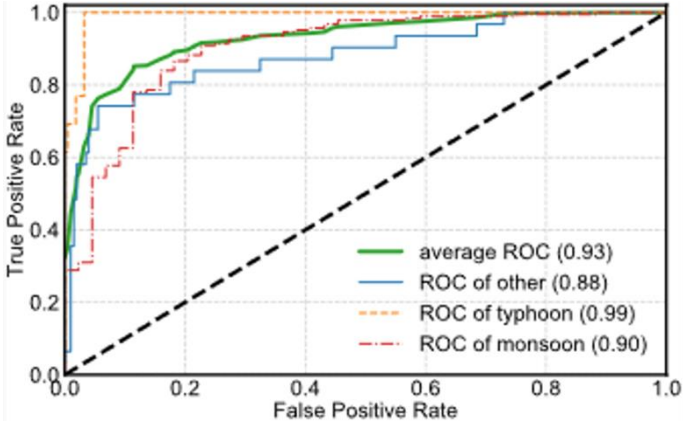

Figure 9. Dachen Island Meteorological Station identification ROC and "average accuracy"

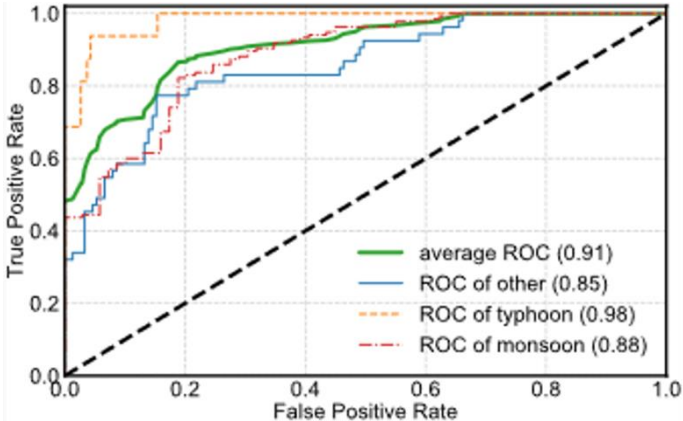

Figure 10. Dinghai Meteorological Station identification ROC and "average accuracy"

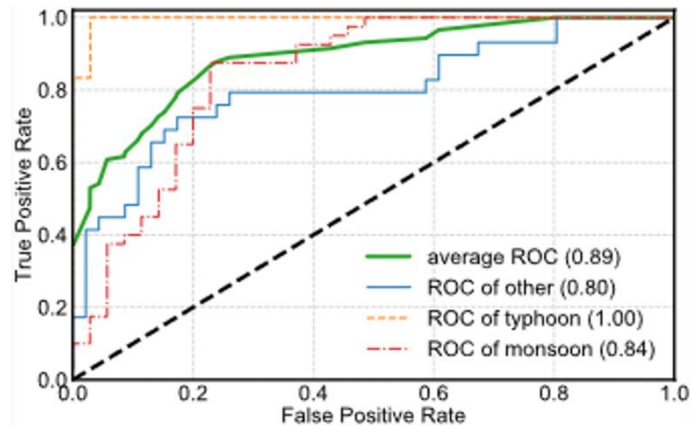

Figure 11. Shengzhou Meteorological Station identification ROC and "average accuracy"

It can be seen from the ROC that the accuracies of the three wind disaster prediction models are within expectations for all three stations, but the typhoon recall in the datasets of the two coastal stations is relatively low. The main reason for this is that several weak typhoons were recorded in the coastal meteorological stations, and their typhoon characteristics were unremarkable for the prediction model. The ROC curve shows that the model has relatively good identification ability for typhoons and monsoons in the three data sets, but relatively poor identification ability for the "other" types of wind disasters, because the sampling frequency of the meteorological data is low (3h/ record) and it is impossible to distinguish more local climatic features. This limitation can be solved by increasing the frequency of data sampling, but currently this method is difficult to popularize.

## 4.5. Performance of cross-station identification model

After verifying that the wind disaster identification model performs well with both training data and testing data from the same station, it is tempting to further prove its scalability. The wind disaster identification model trained by historical data training from one station can be used to identify the wind disaster type for other nearby stations. This task is called cross-station identification of the wind disaster identification algorithm, which will be carried out in this section.

Two different cross-station combinations are implemented: Dinghai-Dachen Island and Dinghai-Shengzhou. For each combination, the data from the former station is used for model training, and the data from the latter station is used for model testing. Table 5 lists the results of the Dinghai-Dachen Island cross-station identification experiment, and Table 6 lists the results of the Dinghai-Shengzhou cross-station identification experiment.

Table 5. Results of Dinghai-Dachen Island cross-station identification experiment

| Training Station Name | Testing Station Name | Types of wind disaster | Precision | Recall | Average Accuracy (AUC) |
|---|---|---|---|---|---|
| Dinghai | Dachen Island | **Typhoon** | 0.731 | 0.743 | |
| | | **monsoon** | 0.895 | 0.695 | **0.795** |
| | | **Other** | 0.565 | 0.796 | |
| Dachen Island | Dinghai | **Typhoon** | 0.703 | 0.778 | |
| | | **monsoon** | 0.900 | 0.931 | **0.850** |
| | | **Other** | 0.663 | 0.503 | |

Table 6. Results of Dinghai-Shengzhou cross-station identification experiment

| Training Station Name | Testing Station Name | Types of wind disaster | Precision | Recall | Average Accuracy (AUC) |
|---|---|---|---|---|---|
| Dinghai | Shengzhou | **Typhoon** | 0.788 | 0.650 | |
| | | **monsoon** | 0.857 | 0.717 | **0.703** |
| | | **Other** | 0.554 | 0.717 | |
| Shengzhou | Dinghai | **Typhoon** | 0.600 | 0.833 | |
| | | **monsoon** | 0.924 | 0.548 | **0.660** |
| | | **Other** | 0.488 | 0.844 | |

The results of the Dinghai-Dachen Island group are better than those of the Dinghai-Shengzhou group in terms of performance index. This is possibly because Dinghai meteorological station and Dachen Island meteorological station are both in the coastal region, and their wind climate environments are very similar. However, Shengzhou meteorological station is more than 70 km from the coast and is considered as an inland meteorological station. Thus, the cross-station identification from Dinghai to Shengzhou is much more challenging.

From comparison of experimental performance (cross-station identification experiment and self-identification experiment), the results from Dinghai-Dachen Island cross-station are 5% less than those from self-identification for Dinghai and Dachen Island station, and the average accuracy of the cross-station identification experiment is about 80%. However, the performance of the Dinghai-Shengzhou group is 10-20% lower than that of the corresponding stations, and its average accuracy is less than 70%, but still above 66%.

In summary, the above results show that the machine learning algorithm for the wind disaster identification model for cross-station identification is good. However, in order to establish a regional wind disaster identification model rather than a single station one, the wind environments of the stations in the region should be relatively consistent.

## 5. Extreme wind speeds analysis for mixed wind climate

Based on identification results from wind disaster classification models, an extreme wind speed for different return periods in a mixed wind climate for structural design is established. Results are compared with those of traditional extreme wind speed estimation methods to verify the influence of mixed climates, which are identified by the proposed model, on the extreme wind speed prediction in mixture climate areas.

### 5.1. Extreme Wind Speed Samples for different wind climate types

Figure 12, Figure 13 and Figure 14 show polar plots of monsoon, typhoon and other wind disaster data obtained from the three stations through the method of construction of extreme samples. The directionalities of the different wind disaster types of the three meteorological

observatories are shown in the figures below.

For Dachen Island, the strong wind speeds of monsoon, typhoon and other disasters mostly come from the northeast-southeast region. Typhoon samples account for the majority of high-speed wind samples, while monsoon and other disasters account for the majority of low-speed wind samples. Comparison of the numbers of high typhoon wind speed records of three stations shows that the typhoon proportions decrease with distance from station to coastline.

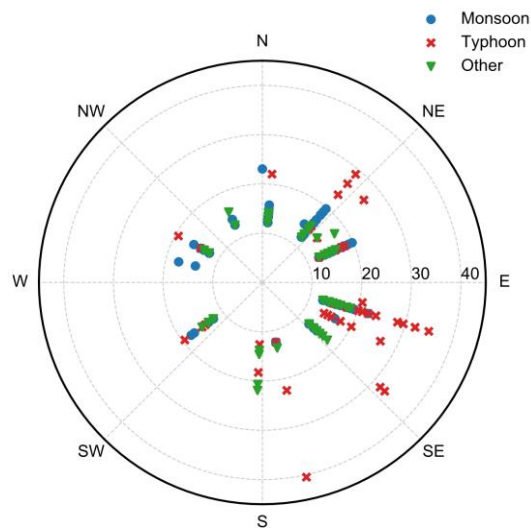

Figure 12. Dachen Island Meteorological Station Extreme Wind Speed Sample Wind Speed-Wind Direction Polar Coordinate Map

## 5.2. Extreme Wind Speed Estimation and Comparison

Extreme wind speed samples of typhoon, monsoon and other wind disasters at Dachen Island, Dinghai and Shengzhou meteorological observatories are constructed using the above methods. Since the filtering threshold is set as 12m/s, except for typhoon wind type, the subsequent calculation of extreme wind speed in this paper is based on the method of peak-over-threshold (POT), i.e. using extreme samples to fit the generalized Pareto distribution (GPD). It should be

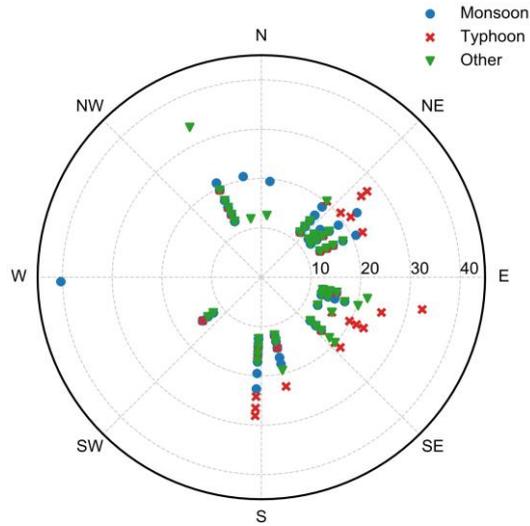

Figure 13. Dinghai Meteorological Station Extreme Wind Speed Sample Wind Speed-Wind

Direction Polar Coordinate Map

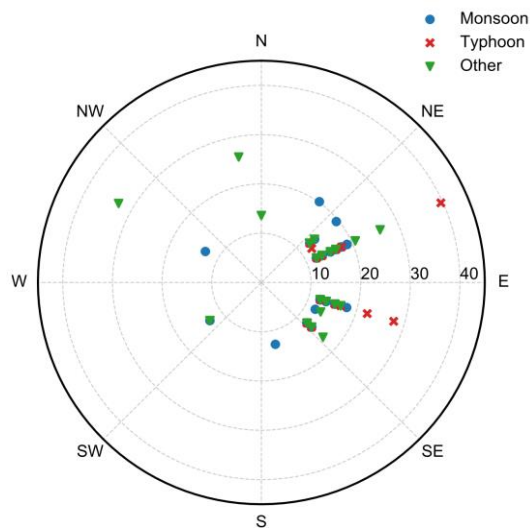

Figure 14. Shengzhou Meteorological Station Extreme Wind Speed Sample Wind Speed-

Wind Direction Polar Coordinate Map

noted that because the wind speed of most sample points in the parent sample of typhoon exceeds the filtering threshold of 12m/s, the Gumbel distribution is used to fit the distribution function.

The extremum wind for mixed wind climates is also identified in this paper, and includes all three wind speed types: typhoons, monsoons and other. Commingled extreme wind speeds are also calculated and compared against China's current building loading code (GB50009-2012). However, from the statistical perspective, a mixture of extreme value samples does not guarantee that the sample points are identically distributed, for example typhoon and monsoon distributions have distinct probabilistic distribution features. Therefore, the extreme value analysis using a mixture of extreme samples is biased for regions influenced by mixed wind climates, such as the three stations analyzed in this paper.

According to Simiu and Yeo (2019), when considering Typhoon, monsoon and other wind disasters separately (Simiu and Yeo 2019), the mixture extreme wind speed cumulative probability distribution function (CDF) is expressed as :

$$P(\max(v_M, v_T, v_O) \leq V) = P(v_M \leq V)P(v_T \leq V)P(v_O \leq V) \tag{5}$$

In the formula, the left side is the CDF of mixture extreme wind speeds, and the right side is the product of CDFs of typhoons, monsoons and other wind disaster speeds. The premise of Eq.(5) is that typhoons, monsoons and other wind disasters are independent of each other.

Figure 15, Figure 16 and Figure 17 show extreme wind speeds for different return periods obtained by observatories in Dachen Island, Dinghai and Shengzhou. The blue solid line and orange dotted line are the monsoon and other extreme wind speed curves calculated using the generalized Pareto distribution through POT samples of monsoon and other. The red dashed line is the typhoon extreme wind speed curve calculated using the generalized Gumbel distribution through typhoon POT samples. The commingled curve is the green dotted line

shown below, which is calculated using the generalized Gumbel distribution through annual maximum samples of all wind speeds. The purple points indicate the extreme wind speeds in the Chinese code, which are almost on the green dotted line. The thick black solid line indicates the extreme wind speed from the mixture distribution above through POT samples of all wind speeds. The curve calculation methods are shown in Table 7.

Table 7. Extreme wind speed for different return period calculation methods

| Types of extreme wind speed | Sample Type | Sample Extraction Method | Distribution |
|---|---|---|---|
| Monsoon | Monsoon | Peak over threshold | Generalized Pareto Distribution |
| Typhoon | Typhoon | Peak over threshold | Generalized Gumbel Distribution |
| Other | Other | Peak over threshold | Generalized Pareto Distribution |
| Commingled | All | Annual maximum | Generalized Gumbel Distribution |
| Mixture | All (classified) | Peak over threshold | Mixture Distribution |

Since the extremum samples are filtered based on the cross-domain method, the relationship between return period and surpass probability is as shown in Formula 6:

$$P(\mathrm{v}_x \geq V) = \frac{1}{T \times N_x} \tag{6}$$

In the formula, x represents the type of wind disaster or extreme wind speed samples, and $N_x$ represents the average number of times such samples occur every year.

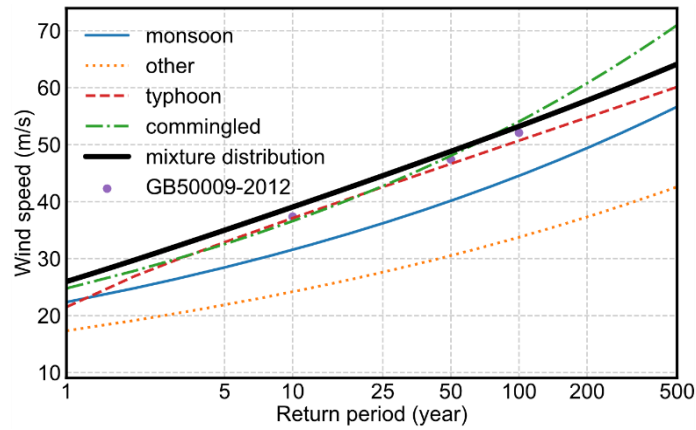

Figure 15. Extreme wind speed-recurrence period curve of Dachen Island Meteorological Station

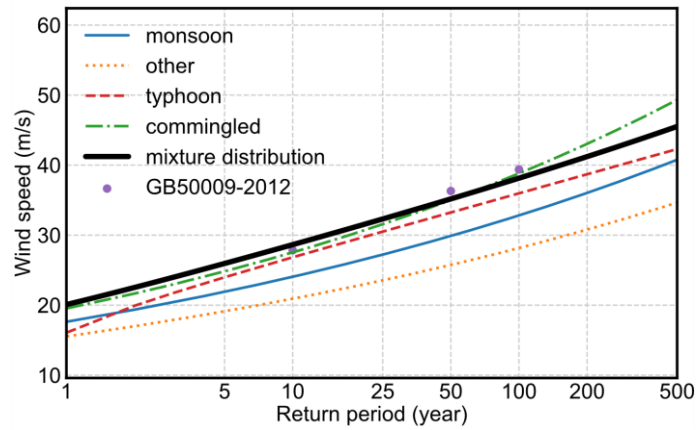

Figure 16. Extreme wind speed-recurrence period curve of Dinghai Meteorological Station

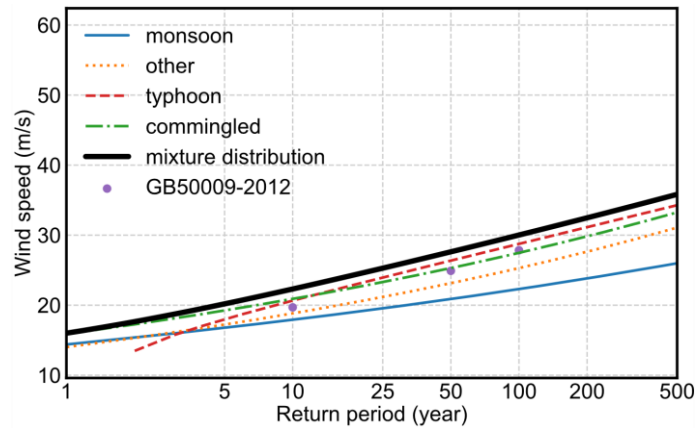

Figure 17. Extreme wind speed-recurrence period curve of Shengzhou Meteorological Station

In this paper, the curves of extreme wind speed - return period for three types of wind disasters are compared. Figure 15, Figure 16 and Figure 17 show that in coastal areas, typhoons control extreme wind speeds with high return periods, while monsoon and other disasters control extreme wind speeds with low return periods. From Figure 15 to Figure 17, the intersection points of the typhoon extreme value wind speed curves and the monsoon extreme value wind speed curves move continuously to the right, which confirms that the demarcation year in which typhoons and monsoons play a controlling role is negatively correlated with distance from the sea, that is, the farther from the sea, the longer the interval of monsoon control. The relationship between monsoon and other wind disasters is also related to distance from the sea. The extreme wind speed at Dachen Island Station is only affected by typhoons and monsoon. The curve distance between monsoon and other wind disasters gets closer in the processes of Figure 15 to Figure 17. The extreme wind speed of other wind disasters at Shengzhou Station exceeds that of monsoon.

The extreme wind speed calculated from the mixture distribution is more consistent with the distribution of samples than the single distribution of mixture extreme samples. The extreme wind speed of mixture distribution is always higher than that of typhoons, monsoon and other disasters in the same return period for different return periods, while the extreme wind speed of mixture extreme value samples is smaller than that of typhoons for some return periods, such as the 5-25 year interval of Dachen Island Station and the 15-year interval of Shengzhou station. This shows that wind speeds for different return periods based on mixture extreme value samples are not always conservative, but a conservative solution of a full return period can be obtained by using the mixture distribution. In addition, the return period of the intersection point of the mixture distribution extreme wind speed curve and the mixture extreme sample extreme wind speed curve is negatively correlated with distance from the sea.

## 6. Conclusion

This paper first proposed the concept of data-driven wind disaster type identification for efficient identification of wind weather data from conventional weather stations that is faster and more accurate than traditional experience-driven classification methods. Based on the concept of wind disaster type identification, this paper compares six common machine learning classification models. Through the K-fold cross-validation method, the most optimal model for wind disaster identification is the support vector machine model (SVM). Secondly, this paper proposed a weather time-course data preprocessing and feature extraction method suitable for wind disaster data classification and developed a wind disaster identification algorithm based on the extracted features. Through calculation using data from three actual meteorological observatories on China's southeast coast, it is shown that the algorithm can realize a historical data training model based on a single station and then identify the type of windstorm corresponding to the new meteorological data of the station. It was also proven that the wind disaster identification algorithm based on the model of data training from one station can identify the wind disaster type corresponding to other stations in the same climate region. Finally, based on the results of wind disaster classification, this paper established separated extreme wind speed samples for different wind disaster types. The difference between the extreme and maximum wind speeds of mixed and unclassified mixed extreme samples was compared, and the mixed distribution obtained a conservative solution for the full return period. Therefore, the wind disaster type has great significance in estimating the extreme wind speed of different return periods, making the structural design performance of the building more reasonable.

**Acknowledgments**

The authors gratefully acknowledge the support of Shanghai Pujiang Program (No. 19PJ1409800), National Key research and Development Program of China (2018YFC0809600, 2018YFC0809604) and National Natural Science Foundation of China (51678451). Any opinions, findings and conclusions or recommendations are those of the authors and do not necessarily reflect the views of the above agencies. Related program can be found at:

https://github.com/cuiwei0322/DataBasedWindDisasterIdentification.